# THE DESIGN OF PARALLEL KINEMATIC MACHINE TOOLS USING KINETOSTATIC PERFORMANCE CRITERIA

Félix Majou, Philippe Wenger, Damien Chablat[*]

## 1. INTRODUCTION

Most industrial machine tools have a serial kinematic architecture, which means that each axis has to carry the following one, including its actuators and joints. High Speed Machining highlights some drawbacks of such architectures: heavy moving parts require from the machine structure high stiffness to limit bending problems that lower the machine accuracy, and limit the dynamic performances of the feed axes.

That is why PKMs attract more and more researchers and companies, because they are claimed to offer several advantages over their serial counterparts, like high structural rigidity and high dynamic capacities. Indeed, the parallel kinematic arrangement of the links provides higher stiffness and lower moving masses that reduce inertia effects. Thus, PKMs have better dynamic performances. However, the design of a parallel kinematic machine tool (PKMT) is a hard task that requires further research studies before wide industrial use can be expected.

Many criteria need to be taken into account in the design of a PKMT. We pay special attention to the description of kinetostatic criteria that rely on the conditioning of the Jacobian matrix of the mechanism. The organisation of this paper is as follows: next section introduces general remarks about PKMs, then is explained why PKMs can be interesting alternative machine tool designs. Then are presented existing PKMTs. An application to the design of a small-scale machine tool prototype developed at IRCCyN is presented at the end of this paper.

[*] Félix Majou, Philippe Wenger, and Damien Chablat, Institut de Recherche en Communications et Cybernétique de Nantes, UMR n°6597 CNRS, École Centrale de Nantes, Université de Nantes, École des Mines de Nantes, 1, rue de la Noë, BP 92101, 44321 Nantes cedex 03 – France.

## 2. ABOUT PARALLEL KINEMATIC MACHINES

### 2.1. General Remarks

The first industrial application of PKMs was the Gough platform (Figure 1), designed in 1957 to test tyres[1]. PKMs have then been used for many years in flight simulators and robotic applications[2] because of their low moving mass and high dynamic performances. Since the development of high speed machining, PKMTs have become interesting alternative machine tool designs[3, 4].

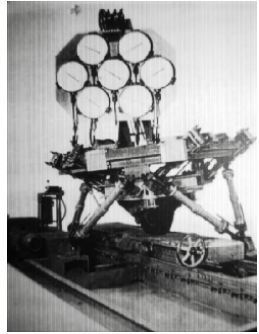

**Figure 1.** The Gough platform

In a PKM, the tool is connected to the base through several kinematic chains or legs that are mounted in parallel. The legs are generally made of either telescopic struts with fixed node points (Figure 2a), or fixed length struts with gliding node points (Figure 2b).

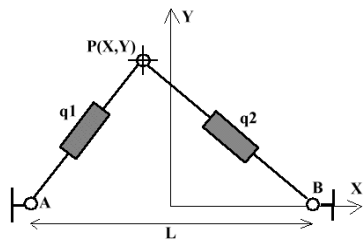 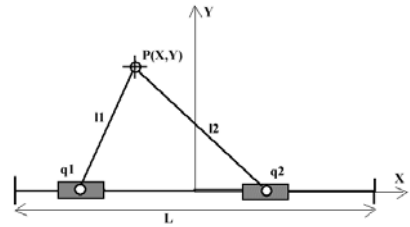

**Figure 2a.** A bipod PKM  **Figure 2b.** A biglide PKM

### 2.2. Singularities

The singular configurations (also called singularities) of a PKM may appear inside the workspace or at its boundaries. There are two types of singularities[5]. A configuration where a finite tool velocity requires infinite joint rates is called a serial singularity. A configuration where the tool cannot resist any effort and in turn, becomes uncontrollable, is called a parallel singularity. Parallel singularities are particularly undesirable because they induce the following problems:
  - a high increase in forces in joints and links, that may damage the structure,
  - a decrease of the mechanism stiffness that can lead to uncontrolled motions of the tool though actuated joints are locked.

Figures 3a and 3b show the singularities for the biglide mechanism of Fig. 2b. In Fig. 3a, we have a serial singularity. The velocity amplification factor along the vertical direction is null and the force amplification factor is infinite.

Figure 3b shows a parallel singularity. The velocity amplification factor is infinite along the vertical direction and the force amplification factor is close to zero. Note that a high velocity amplification factor is not necessarily desirable because the actuator encoder resolution is amplified and thus the accuracy is lower.

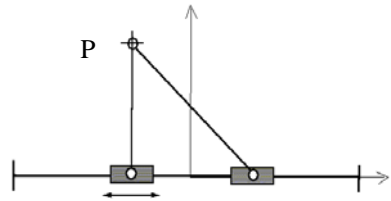
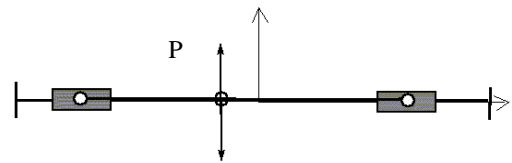

**Figure 3a.** A serial singularity        **Figure 3b.** A parallel singularity

## 2.3. Working and Assembly Modes

A serial (resp. parallel) singularity is associated with a change of working mode[6] (resp. of assembly mode). For example, the biglide has four possible working modes for a given tool position (each leg node point can be to the left or to the right of the intermediate position corresponding to the serial singularity, Fig. 4a) and two assembly modes for a given actuator joint input (the tool is above or below the horizontal line corresponding to the parallel singularity, Fig. 4b). The choice of the assembly mode and of the working mode may influence significantly the behaviour of the mechanism[5].

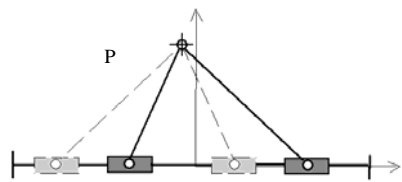
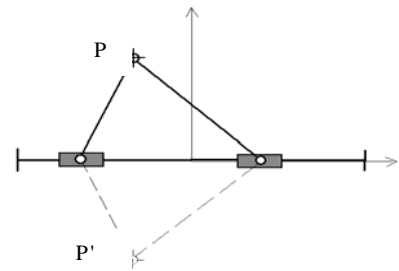

**Figure 4a.** The four working modes        **Figure 4b.** The two assembly modes

## 3. PKMs AS ALTERNATIVE MACHINE TOOL DESIGNS

### 3.1. Limitations of Serial Machine Tools

Today, newly designed machine tools benefit from technological improvements of components such as spindles, linear actuators, bearings. Most machine tools are based on a serial architecture (Figure 5), whose advantage is that input/output relations are simple.

Nevertheless, heavy masses to be carried and moved by each axis limit the dynamic performances, like feed rates or accelerations. That is why machine tools manufacturers have started being interested into PKMs since 1990.

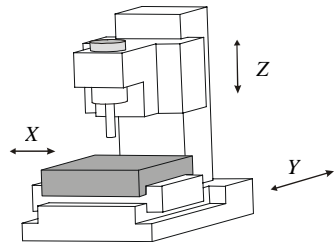

**Figure 5.** A serial PPP machine tool

### 3.2. PKMs Potentialities for Machine Tool Design

The low moving mass of PKMs and their good stiffness allow high feed rates (up to 100 m/min) and accelerations (from 1 to 5g), which are the performances required by High Speed Machining.

PKMs are said to be very accurate, which is not true in every case[4], but another advantage is that the struts only work in traction or compression. However, there are many structural differences between serial and parallel machine tools, which makes it hard to strictly compare their performances.

### 3.3. Problems with PKMs

a) The workspace of a PKM has not a simple geometric shape, and its functional volume is reduced, compared to the space occupied by the machine[7], as we can see on Fig. 6.

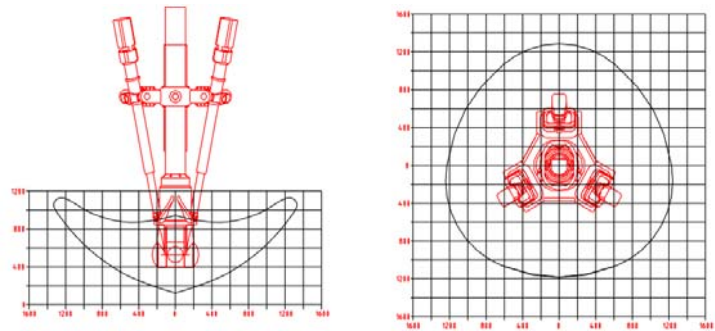

**Figure 6.** Workspace sections of Tricept 805 (www.neosrobotics.com)

b) For a serial mechanism, the velocity and force transmission ratios are constant in the workspace. For a parallel mechanism, in contrast, these ratios may vary significantly in the workspace because the displacement of the tool is not linearly related to the displacement of the actuators. In some parts of the workspace, the maximal velocities and forces measured at the tool may differ significantly from the maximal velocities and forces that the actuators can produce. This is particularly true in the vicinity of singularities. At a singularity, the velocity, accuracy and force ratios reach extreme values.

c) Calibration of PKMs is quite complicated because of kinematic models complexity[8].

## 4. EXISTING PKMT DESIGNS

In this section will be presented some existing PKMTs.

### 4.1. Fully Parallel Machine Tools

What we call fully parallel machine tools are PKMs that have as many degrees of freedom as struts. On Fig. 7, we can see a 3-*RPR* fully parallel mechanism with three struts. Each strut is made of a revolute joint, a prismatic actuated joint and a revolute joint.

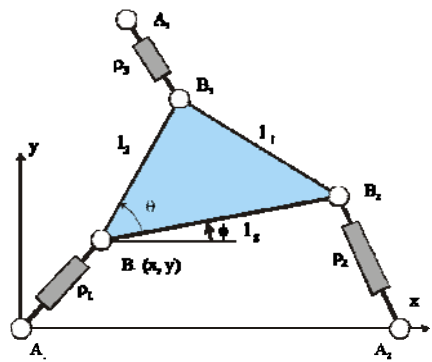

**Figure 7.** 3-*RPR* fully parallel mechanism

Fully PKMT with six variable length struts are called hexapods. Hexapods are inspired by the Gough Platform. The first PKMT was the hexapod "Variax" from Giddings and Lewis presented in 1994 at the IMTS in Chicago. Hexapods have six degrees of freedom. One more recent example is the CMW300, a hexapod head designed by the Compagnie Mécanique des Vosges (Figure 8).

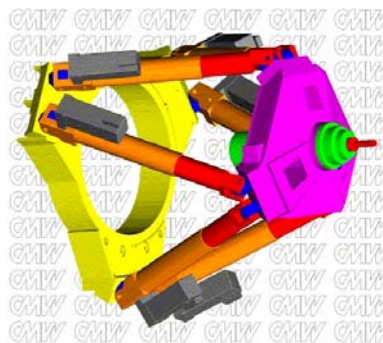

**Figure 8.** Hexapod CMW 300 (perso.wanadoo.fr/cmw.meca.6x/6AXES.htm)

Fully parallel machine tools with fixed length struts can have three, four or six legs. The Urane SX (Figures 9 and 14) from Renault Automation is a three leg machine, whose tool can only move along X, Y and Z axes, and its architecture is inspired from the Delta robot[9], designed for pick and place applications. The Hexa M from Toyoda is a PKMT with six fixed length struts (Figure 10).

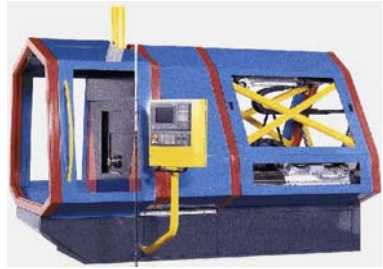 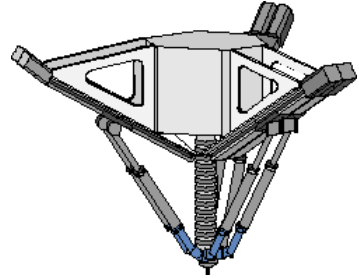

**Figure 9.** Renault automation Urane SX (from "Renault Automation Magazine", n° 21, may 1999)

**Figure 10.** Toyoda Hexa M (www.toyoda-kouki.co.jp)

### 4.2. Other Kinds of PKMT

The Tricept 805 is a widely used PKMT with three variable length struts (Figures 6 and 11). The Tricept 805 has a hybrid architecture: a two degrees of freedom wrist serially mounted on a tripod architecture.

Another non fully parallel MT is the Eclipse (Figure 12) from Sena Technology[10, 11]. The Eclipse is an overactuated PKM for rapid machining, capable of simultaneous five faces milling, as well as turning, thanks to the second spindle.

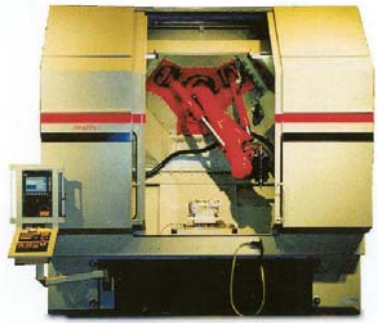 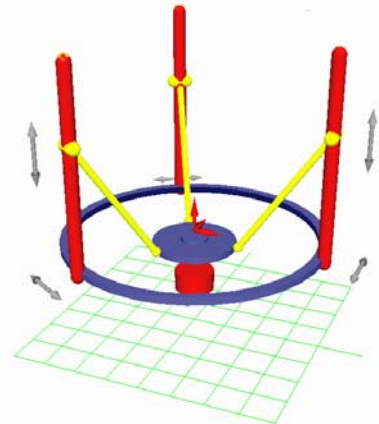

**Figure 11.** Tricept 805 from Neos robotics (www.neosrobotics.com)

**Figure 12.** The Eclipse, from Sena Technology (macea.snu.ac.kr/eclipse/homepage.html)

## 5. DESIGNING A PKMT

### 5.1. A Global Task

Given a set of needs, the most adequate machine will be designed through a set of design parameters like the machine morphology (serial, parallel or hybrid kinematic structure), the machine geometry (link dimensions, joint orientation and joint ranges), the type of actuators (linear or rotative motor), the type of joints (prismatic or revolute), the number and the type of degrees of freedom, the task for which the machine is designed. These parameters must be defined using relevant design criteria.

### 5.2. Kinetostatic Performance Criteria are Adequate for the Design of PKMTs

The only way to cope with problems due to singularities is to integrate kinetostatic performance criteria in the design process of a PKMT. Kinetostatic performance criteria evaluate the ability of a mechanism to transmit forces or velocities from the actuators to the tool. These kinetostatic performance criteria must be able to guaranty minimum stiffness, accuracy and velocity performances along every direction throughout the workspace of the PKMT.

To reach this goal, we use two complementary criteria: the conditioning of the Jacobian matrix **J** of the PKMT, called conditioning index, and the manipulability ellipsoid associated with **J**[12]. The Jacobian matrix **J** relates the joint rates to the tool velocities. It also relates the static tool efforts to the actuator efforts. The conditioning index is defined as the ratio between the highest and the smallest eigenvalue of **J**. The conditioning index varies from 1 to infinity. At a singularity, the index is infinity. It is 1 at another special configuration called isotropic configuration. At this configuration, the tool velocity and stiffness are equal in all directions. The conditioning index measures the uniformity of the distribution of the velocities and efforts around one given configuration but it does not inform about the magnitude of the velocity amplification or effort factors.

The manipulability ellipsoid is defined from the matrix $(\mathbf{J}\mathbf{J}^T)^{-1}$. The principal axes of the ellipsoid are defined by the eigenvectors of $(\mathbf{J}\mathbf{J}^T)^{-1}$ and the lengths of the principal axes are the square roots of the eigenvalues of $(\mathbf{J}\mathbf{J}^T)^{-1}$. The eigenvalues are associated with the velocity (or force) amplification factors along the principal axes of the manipulability ellipsoid.

These criteria are used in Wenger[13], to optimize the workspace shape and the performances uniformity of the Orthoglide, a three degree of freedom PKM dedicated to milling applications (Figure 13).

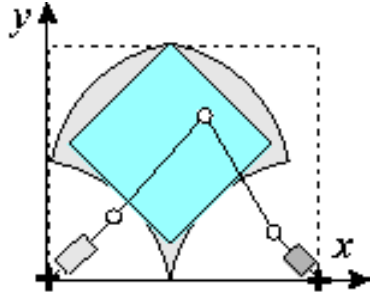

**Figure 13.** A section of Orthoglide's optimised workspace

### 5.3. Technical Problems

If the struts of the PKMT are made with ballscrews, the PKMT accuracy may suffer from struts warping due to heating caused by frictions generated by ballscrews. This problem is met by hexapods designers that use ballscrews. Thus, besides manufacturing inaccuracies, the calibration of a PKMT will have to take into account dimensions variations due to dilatation. A good thermal evacuation can minimise the effects of heating.

In case PKMT actuators are linear actuators, magnetic pollution has to be taken into account so that chips clearing out is not obstructed. One technique, used by Renault Automation for the Urane SX, is to isolate the tool from the mechanism (Figure 14).

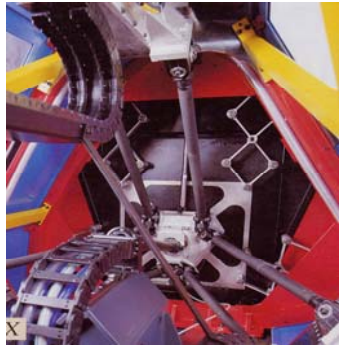

**Figure 14.** Urane SX tool is isolated from the mechanism that generates the motion (from "Renault Automation Magazine", n° 21, may 1999)

At last, choosing fixed length or variable length struts influence the behaviour of the machine. Actuators have to be mounted on the struts in case of variable length struts, which increases moved masses. Fixed length struts do not have this problem, and furthermore allow the use of linear actuators, that bring high dynamic performances.

### 6. THE ORTHOGLIDE

A three-axis PKMT, the Orthoglide, was designed at IRCCyN Using the two kinetostatic criteria given in section 5.2. A small-scale prototype is under development (Figure 15).

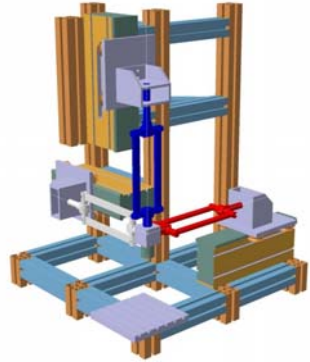

**Figure 15.** PKMT prototype

The theoretical kinematic analysis of the Orthoglide is described in Wenger[13]. The Orthoglide has been designed such that it has an isotropic configuration (the velocity and forces transmission factors are equal in each direction) in its workspace. Moreover the velocity and force transmission factors remain under prescribed values throughout the workspace. The paper describes how the kinetostatic criteria have been used to define the Orthoglide kinematics and geometry for machine tool applications (Figure 16).

The dimensions of the Orthoglide have been chosen for a 200*200*200 prescribed cartesian workspace, so that the velocity and force amplification factors remain between 0.6 and 1.7, which guarantees minimum performances inside the available workspace. Joint ranges have also been optimised to respect the limits chosen for the velocity and force transmission factors (Figure 17).

One or two degrees of freedoms can be added to the Orthoglide either by:

- mounting serially a one or two degree of freedom wrist, or
- mounting an orienting branch on the base of the Orthoglide, leading to a left hand / right hand architecture.

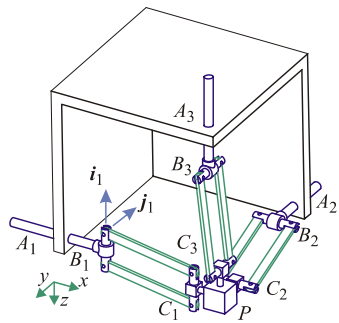         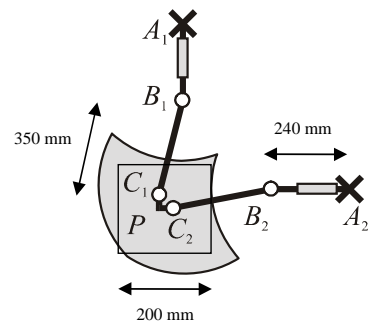

**Figure 16.** Kinematics of the Orthoglide        **Figure 17.** Dimensions of the Orthoglide

## 7. CONCLUSIONS

The aim of this article was to introduce a few criteria for the design of PKMTs, which may become interesting alternatives for High Speed Machining, especially in the milling of large parts made of hard material, or for serial manufacturing operations on aeronautical parts.

Kinetostatic criteria seem to be well adapted to the design of PKMTs, particularly for the kinematic design and for the optimisation of the workspace shape, with regard to performances uniformity.

The kinetostatic criteria have been used for the design of the Orthoglide, a three-axis PKMT developed at IRCCyN. A small scale prototype is under development. A five-axis PKMT will be derived from the Orthoglide.